\documentclass{article}

    \PassOptionsToPackage{numbers, compress}{natbib}

 \usepackage[main, final]{neurips_2026}

\usepackage[utf8]{inputenc} 
\usepackage[T1]{fontenc}    
\usepackage{hyperref}       
\usepackage{url}            
\usepackage{booktabs}       
\usepackage{amsfonts}       
\usepackage{nicefrac}       
\usepackage{microtype}      

\usepackage{pifont}
\usepackage{wrapfig}
\usepackage{amsmath,amsfonts,amssymb,amstext,bm}
\usepackage{bbm}
\usepackage{multirow}
\usepackage[table]{xcolor}
\usepackage{xcolor}         
\usepackage{subcaption}
\usepackage{dashrule}
\usepackage[ruled]{algorithm2e}
\usepackage{lipsum}
\usepackage{tcolorbox}
\usepackage{pgfplots}
\usetikzlibrary{patterns,backgrounds}
\usepgfplotslibrary{groupplots}
\definecolor{r_d}{RGB}{162, 0, 037}
\definecolor{r_l}{RGB}{248, 206, 204}
\definecolor{g_d}{RGB}{0, 87, 0}
\definecolor{g_l}{RGB}{213, 232, 212}
\definecolor{b_d}{RGB}{0, 110, 175}
\definecolor{b_l}{RGB}{218, 232, 252}
\definecolor{o_d}{RGB}{180, 101, 4}
\definecolor{o_l}{RGB}{255, 230, 204}
\title{Simple Test-Time Refinement for Plot-to-Code Generation via Visual-Code Diagnostics}

\author{%
  Peiming Guo\footnotemark[2], Jinpeng Huang\footnotemark[2], Yangbin Sun\footnotemark[2], Gaoxiong Cao\footnotemark[3], \\ \textbf{Ensheng Shi\footnotemark[3], Yuchi Ma\footnotemark[3], Yu Zhang\footnotemark[4], Meishan Zhang\footnotemark[2]~\thanks{~~Corresponding author: Meishan Zhang.}, Baotian Hu\footnotemark[2], Min Zhang\footnotemark[2]}  \\
\footnotemark[2]~~Institute of Computing and Intelligence, Harbin Institute of Technology (Shenzhen), \\ \footnotemark[3]~~Huawei CodeArts Model Team, HUAWEI, \\ \footnotemark[4]~~School of Computer Science and Technology, Soochow University \\
\texttt{guopeiming.gpm@gmail.com, mason.zms@gmail.com}
}

\begin{document}

\maketitle

\begin{abstract}
Plot-to-code generation has been greatly advanced by recent multimodal large language models (MLLMs). However, existing studies mainly focus on one-pass generation or iterative sampling guided by verifier-based filtering, which does not align with the typical human programming workflow of generate first, then iteratively error-location and fix. This naturally motivates a test-time refinement paradigm for plot-to-code generation.
In this work, we present the first study on plot-to-code generation via iterative test-time refinement. We propose a Visual-Code Diagnostics framework that enables refinement through visual feedback from source code. Specifically, we first categorize generation errors into four coarse-grained types, and then train lightweight discriminators to recognize and localize these errors based on the rendered visual outputs and corresponding code. Unlike traditional compiler- or debugger-based verification, our approach performs iterative refinement guided by one diagnostics system, enabling more effective semantic and visual correction during test time.
Extensive experiments demonstrate that our method consistently improves generation quality across diverse generators and significantly outperforms strong test-time scaling baselines. 

\end{abstract}

\section{Introduction}

Plot-to-code generation~\citep{yang2025chartmimic,wu-etal-2025-plot2code,zhao-etal-2025-chartcoder,tan2025chartmaster,niu-etal-2025-chart2code53} aims to synthesize executable plotting programs that faithfully reproduce a given reference chart. 
This task has attracted growing attention due to its applications in visual chart understanding~\citep{xu2025chartmoe,Yang_2025_ICCV}, data visualization~\citep{yang2025chartmimic,ford-etal-2025-charting}, and code generation~\citep{duan2025codeplot,masry-etal-2025-chartgemma,carbune-etal-2024-chart}. 
Recent advances in large multimodal language models (MLLMs)~\citep{qwen3vl,hurst2024gpt,yao2025efficient,wang2025internvl3,hong2025glm,guo2025seed1,wu2024deepseek} have substantially improved performance by enabling joint visual and textual reasoning. 
Nevertheless, generating visually accurate plotting code remains challenging, especially for complex charts requiring precise structural layouts, data mappings, and stylistic consistency.

Most existing approaches~\citep{zhang2025boosting,tan2025chartmaster,chen2026breaking, zhao-etal-2025-chartcoder, niu-etal-2025-chart2code53} formulate plot-to-code generation as a direct program synthesis problem, where the model generates plotting code from the input chart in a single pass. 
Although recent MLLMs~\citep{team2025kimi,10445007,wu2024deepseek,huang2025visual,cao2025hunyuanimage,11208999} demonstrate strong generation capabilities, such methods remain highly sensitive to visual-semantic errors, including incorrect data encodings, missing visual elements, layout and styling mismatches. 
In practice, we observe that these failures exhibit recurring visual patterns and can be grouped into several coarse-grained categories of visual errors. 
Once such errors appear in generated code, they are often difficult to correct through single-pass decoding alone.

To alleviate generation failures, recent works have begun exploring test-time scaling strategies such as iterative repeated sampling~\citep{brown2024large}. 
Repeated sampling methods improve performance by generating multiple candidate programs and selecting outputs using execution-based or verifier-based filtering. 
However, these approaches repeatedly regenerate the entire program, leading to substantial inference cost and token consumption, especially for long and structurally complex code. 
More importantly, existing verification signals are typically coarse-grained, only indicating whether a generated result is acceptable without identifying the underlying causes of visual discrepancies. 
Such paradigms do not align with the typical human programming workflow, where developers usually generate an initial code first, then iteratively inspect outputs, localize errors, and revise only problematic code regions. 
This naturally motivates a test-time refinement paradigm for plot-to-code generation.

Iterative refinement provides a more promising direction by progressively correcting generation errors through multiple rounds of revision~\citep{li-etal-2025-metal}. 
However, effective refinement for plot-to-code generation requires more than holistic correctness evaluation. 
Existing approaches still lack fine-grained visual-semantic diagnostics that can explicitly connect rendering errors with suspicious code regions. 
As a result, refinement often remains inefficient, requiring redundant regeneration and suffering from accumulated error propagation. 
This limitation becomes particularly pronounced in multi-agent refinement settings, where the absence of structured diagnostic feedback leads to ineffective interactions between generation and revision agents.

\begin{wrapfigure}{r}{0.65\linewidth}
\centering
\includegraphics[width=0.65\textwidth]{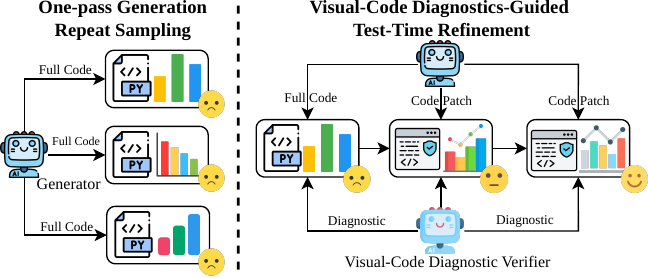}
\caption{Comparison of one-pass generation-based repeat sampling and our visual-code diagnostic-guided test-time refinement.}
\label{fig:intro}
\end{wrapfigure}

To address these challenges, we propose a Visual-Code Diagnostics framework for efficient test-time refinement in plot-to-code generation. 
As shown in \figureautorefname~\ref{fig:intro}, instead of repeatedly regenerating the full plotting program, our framework iteratively diagnoses visual-code inconsistencies from rendered charts and performs localized code patch refinement. 
Specifically, we introduce a lightweight visual-code diagnostic verifier that identifies visual discrepancy categories and localizes suspicious code regions associated with rendering errors. 
The resulting diagnostic feedback serves as a structured interaction bridge for iterative refinement, enabling targeted code modification with substantially lower token consumption and improved correction efficiency.

To train the verifier, we construct a visual-code error discrimination dataset using controlled code mutations generated by MLLMs to simulate realistic plotting errors. 
We further employ an improved line-level Levenshtein distance algorithm~\citep{lcvenshtcin1966binary,10.1145/321796.321811} to localize modified code regions and build structured supervision signals that associate visual discrepancies with corresponding code edits, enabling the verifier to learn visual-semantic diagnostic reasoning over chart-code pairs.
At inference time, our framework follows an iterative refinement pipeline: generate, render, diagnose, and patch-refine. 
Unlike conventional test-time scaling approaches that repeatedly regenerate complete programs, our method only revises localized suspicious code segments during refinement, significantly improving inference efficiency while maintaining strong correction capability.

We evaluate our approach on Chart2Code-160k~\citep{zhao-etal-2025-chartcoder}, ChartMimic~\citep{yang2025chartmimic}, and Plot2Code ~\citep{wu-etal-2025-plot2code}. 
Experimental results demonstrate that our method consistently improves generation quality across multiple generators and substantially outperforms strong test-time scaling baselines. 
Furthermore, compared with existing iterative refinement methods, our framework achieves superior performance with considerably lower inference cost.

Our contributions can be summarized as follows:
\begin{itemize}
    \item We present a simple test-time refinement for plot-to-code generation, creating a diagnostic system to guide the iterative refinement while reducing inference cost.
    \item We build a lightweight visual-code diagnostic verifier that is able to detect visual discrepancies and localize them to specific code regions based on an automatically built dataset.
    \item We show that our method is effective on plot-to-code benchmarks, outperforming strong baselines, including one-pass generation and other test-time scaling strategies at low cost. \footnote{Our code is publicly available at \url{https://github.com/guopeiming/plot_to_code_refine}.}
\end{itemize}

\section{Related Work}

\paragraph{Plot-to-Code Generation.}
With the rapid development of multimodal large language models (MLLMs)~\citep{team2025kimi,10445007,wu2024deepseek,huang2025visual,cao2025hunyuanimage}, plot-to-code generation has attracted increasing attention.
Recent works have constructed various benchmarks and training datasets for this task, including ChartMimic~\citep{yang2025chartmimic}, Plot2Code~\citep{wu-etal-2025-plot2code}, ChartEdit~\citep{zhao-etal-2025-chartedit}, ChartX~\citep{11202357}, ReChartPrompt-240K~\citep{tan2025chartmaster}, Chart2Code-160K~\citep{zhao-etal-2025-chartcoder}, and Chart2Code53~\citep{niu-etal-2025-chart2code53}.
VisCodex~\citep{jiang2025viscodex} and JanusCode~\citep{sun2025januscoder} further extend multimodal code generation to broader tasks such as webpage and UI generation.

Beyond dataset construction, researchers have explored different training paradigms for improving plot-to-code generation.
\citet{zhang2025boosting} optimize generation using preference learning and DPO, while ChartMaster~\citep{tan2025chartmaster} and MSRL~\citep{chen2026breaking} employ reinforcement learning with visual and attribute-level rewards.
ChartCoder~\citep{zhao-etal-2025-chartcoder} and Chart2Code53~\citep{niu-etal-2025-chart2code53} perform supervised fine-tuning on large-scale chart-code datasets.
Most existing methods formulate plot-to-code as direct program synthesis, where the entire plotting program is generated in a single pass.
In contrast, our work focuses on efficient test-time refinement through visual-code diagnostics.

\paragraph{Test-Time Scaling and Iterative Refinement.}
Test-time scaling~\citep{brown2024large,snell2024scaling} has recently become an effective strategy for improving code generation performance.
A common approach is repeated sampling~\citep{brown2024large}, where multiple candidate programs are generated and filtered using execution feedback, reward models, or verifier-based selection~\citep{lightman2024lets}.
However, such methods repeatedly regenerate complete programs, resulting in high inference cost and suboptimal results.

Recent works have also explored iterative and multi-agent refinement for plot-co-code generation.
METAL~\citep{li-etal-2025-metal} introduces a multi-agent framework for chart-to-code generation with iterative collaboration between multiple GPT-4o-based agents.
While effective, existing refinement frameworks still rely on holistic conversational feedback and repeated full-program generation, making them vulnerable to inefficient revisions and error propagation.

Different from prior work, our framework performs localized patch-based refinement through visual-code diagnostics.
By identifying visual discrepancy categories and suspicious code regions, our method enables targeted iterative correction with substantially improved inference efficiency.

\section{Method}

Given a reference chart image $I$, plot-to-code generation aims to synthesize an executable plotting program $C$ using an MLLM generator $D$, such that the rendered chart $\widehat{I}$ faithfully matches $I$.
Instead of treating this task as one-pass generation, we propose the test-time refinement strategy, where an initially generated program is iteratively corrected according to visual-code diagnostic feedback.
Specifically, we introduce visual-code diagnostic verification: $V(I,\widehat{I},C) \rightarrow (e, L)$, 
where $e$ denotes the visual error category and $L$ denotes the suspicious code regions associated with the discrepancy.
Based on $(e,L)$, the generator performs localized patch-based refinement instead of full-program regeneration, enabling more efficient and targeted correction.

\begin{figure*}[th]
    \centering
    \includegraphics[width=1\textwidth]{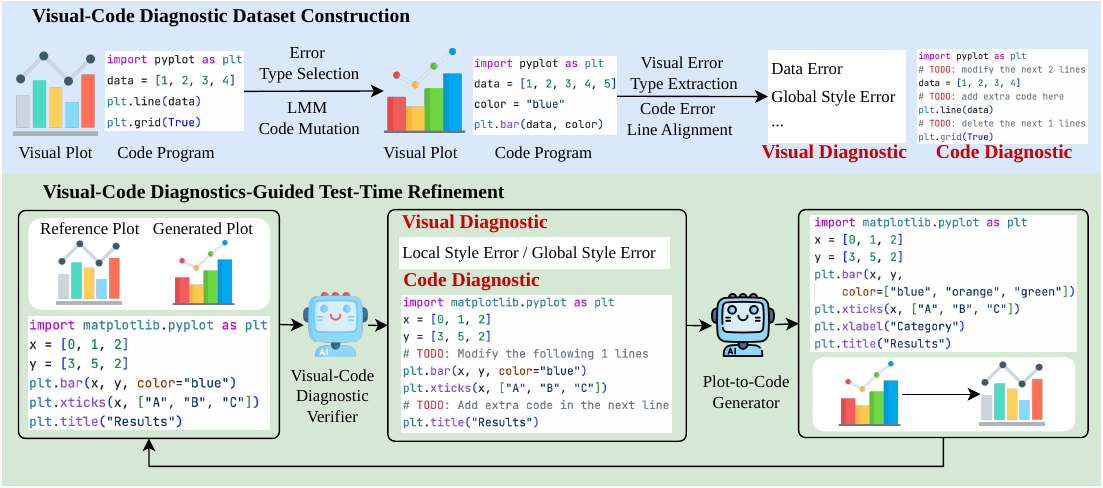}
    \caption{Overview of test-time refinement for plot-to-code generation via visual-code diagnostics.}
    \label{fig:method}
\end{figure*}

This section introduces our framework (\figureautorefname~\ref{fig:method}), including visual error taxonomy and dataset construction, diagnostic verifier training, and visual-code diagnostics-guided test-time refinement.

\subsection{Visual-Code Diagnostic Verification}
We model visual-code diagnostics as structured errors that can be detected from rendered outputs and localized to specific code segments. 
This subsection describes how we define error structures, construct supervision, and train the verifier.

\subsubsection{Error Modeling and Diagnostic Categories}

To better understand failure patterns in plot-to-code generation, we conduct a systematic analysis of real-world errors produced by several representative MLLMs, including ChartMaster-7B~\citep{tan2025chartmaster}, Qwen-VL-2B-Instruct, Qwen-VL-32B-Instruct~\citep{qwen3vl}, and GPT-4o~\citep{hurst2024gpt}. 
Our analysis shows that visual discrepancies typically exhibit structured and localized patterns, rather than arbitrary mismatches. 
Instead of defining a rigid taxonomy, we adopt a set of coarse-grained diagnostic categories that capture common failure modes in Matplotlib-based~\citep{Hunter:2007,harris2020array} plot generation. 
These categories broadly include: 1). compiler failures that prevent successful rendering, 2). data-related error in plotted shapes or values, 3). local styling errors affecting specific visual elements, and 4). global layout or styling inconsistencies involving plot-level components.

Importantly, these errors are often associated with localized code regions, suggesting that effective refinement requires not only detecting whether a chart is incorrect, but also identifying where the error likely originates in the code. 
This observation motivates our diagnostic verifier, which produces structured feedback grounded in code structure and visual discrepancies.
Detailed error definitions, examples, and distribution statistics of real-world errors are provided in \appendixautorefname~\ref{sec:app:error_taxonomy}.

\subsubsection{Structured Diagnostic Dataset Construction}

To train the visual-code diagnostic verifier, we construct a structured dataset that provides supervision for both visual error detection and code localization. Each sample consists of a generated plot, a reference plot, and the code that produces the generated plot, with supervision signals including the visual error type and code-grounded diagnostic annotations.

We build this dataset from correct chart-code pairs in Chart2Code-160K~\citep{zhao-etal-2025-chartcoder}. To simulate realistic errors, we perform error-aware code mutation using GPT-4o~\citep{hurst2024gpt}. Specifically, for each seed program, we generate three mutated variants corresponding to data error, local style error, and global style error, while keeping other parts of the code unchanged. Compiler errors are excluded from mutation, as they can be directly identified via execution feedback. To improve diversity, mutation types are applied in random sequential orders across samples.

Given one seed program and its mutated variants, we construct pairwise diagnostic samples by selecting two programs $(C_1, C_2)$, where $C_1$ is treated as the generated code and $C_2$ as the reference. The corresponding plots $(I_1, I_2)$ are rendered, and the visual error label is determined by the mutation type that differentiates $C_1$ from $C_2$. This pairwise construction yields diverse visual-code discrepancies and expands the dataset size (up to 12$\times$ per seed example).

To obtain code-region supervision, we introduce a line-level alignment algorithm based on an improved Levenshtein distance algorithm~\citep{lcvenshtcin1966binary}. The method identifies inserted, deleted, and modified code segments between $C_1$ and $C_2$, and maps them to candidate error regions.
The algorithm details are placed in \appendixautorefname~\ref{sec:app:dataset:algo}.
Instead of predicting explicit line indices, we represent localization signals by inserting \texttt{TODO} comments into the code near the identified regions, forming structured diagnostic annotations that are more robust for the generation paradigm of MLLMs.

This process converts synthetic visual-code differences into structured supervision and enables the scalable construction of structured diagnostic supervision without manual annotation, allowing the verifier to learn both error classification and fine-grained code localization. Detailed construction procedures are provided in Appendix~\ref{sec:app:dataset_construction}.

\subsubsection{Visual-Code Diagnostic Verifier Training}
\label{sec:method:training}
We train a compact visual-code diagnostic verifier that takes the reference chart $I$, rendered chart $\widehat{I}$, and generated code $C$ as input, and outputs structured visual-code diagnostics $(e, L)$, where visual diagnostic $e$ is the visual error type from our taxonomy and code diagnostic $L$ is localized code feedback.
Specifically, instead of predicting line indices, it generates the code with \texttt{TODO} comments inserted near erroneous regions, providing generation-friendly and context-aware localization.

We train the verifier using a combination of supervised fine-tuning and reinforcement learning. 
While supervised training teaches the model the basic diagnostic format, reinforcement learning further improves the accuracy and usefulness of the diagnostic outputs. 
To this end, we design a reward function for GRPO training~\citep{guo2025deepseek} that evaluates both visual error prediction and code localization quality: $r = r_{\text{visual}} + r_{\text{code}}$.
Both components are computed using Jaccard similarity~\citep{jaccard1901etude}: $r(X, Y) = \frac{|X \cap Y|}{|X \cup Y|},$
where $X$ and $Y$ denote the predicted and ground-truth sets, respectively.

For visual diagnostics, $r_{\text{visual}}$ measures the overlap between predicted and ground-truth error types. 
For code diagnostics, we represent each predicted diagnostic action as a tuple $(i, a, j)$, where $i$ denotes the line position of the inserted \texttt{TODO} comment, $a \in \{\texttt{add}, \texttt{modify}, \texttt{delete}\}$ denotes the repair action type, and $j$ denotes the number of affected lines. 
The reward $r_{\text{code}}$ is computed as the Jaccard similarity between predicted and ground-truth sets of such tuples.

This training formulation aligns the verifier with its diagnostic role: it encourages accurate identification of visual discrepancies while producing precise, code-grounded localization signals. Notably, a compact 2B model is sufficient to perform joint visual-code diagnostics, eliminating the need for complex multi-agent reasoning while enabling efficient test-time refinement with significantly reduced inference cost.

\subsection{Visual-Code Diagnostics-Guided Test-Time Refinement}

Given a reference chart $I$, the generator $D$ first produces an initial program $C_0$, which is rendered into $\widehat{I}_0$. 
The diagnostic verifier then analyzes the visual discrepancy between $I$ and $\widehat{I}_t$, together with the associated code $C_t$, and produces structured diagnostic feedback $(e_t, L_t)$, where $e_t$ denotes a coarse-grained error category and $L_t$ indicates the suspicious code region.

Based on this feedback, the generator produces a localized code patch $\Delta C_t$ targeting the identified region $L_t$. 
The patch is then merged back into the original program to obtain the refined code:
\begin{equation}
    (e_t, L_t) = V(I, \widehat{I}_t, C_t), 
    \qquad
    \Delta C_t = D(C_t, e_t, L_t),
    \qquad
    C_{t+1} = \texttt{Merge}(C_t, \Delta C_t).
\end{equation}

This process can be iteratively repeated to progressively improve chart correctness. 
Unlike one-pass regeneration methods, our approach performs localized refinement while preserving correct code regions, reducing unnecessary modifications and mitigating error propagation. 
Moreover, compared to multi-agent refinement frameworks that require repeated interactions among multiple large models, our method relies on a lightweight diagnostic verifier to provide explicit and structured feedback, resulting in a substantially more efficient test-time refinement process.

\section{Experiments}
We conduct experiments to verify the effectiveness and efficiency of our proposed visual-code diagnostic verifier and test-time refinement strategy.
In addition, we also perform experimental analyses to gain a deeper understanding of plot code diagnostics and repair.

\begin{table*}[th]
\centering

\caption{Main results of plot-to-code generation across different generators, visual-code diagnostic verifiers, and test-time scaling methods. Vanilla represents one-pass generation.}
\resizebox{1\textwidth}{!}{
\begin{tabular}{l|ccccccc|ccccccc}

\toprule
\multirow{2}{*}{\textbf{Models}} & \multicolumn{7}{c|}{\textbf{ChartMimic}} & \multicolumn{7}{c}{\textbf{Plot2Code}} \\ \cmidrule{2-15} 
& \textbf{E} & \textbf{TT} & \textbf{LO} & \textbf{TP} & \textbf{CL} & \textbf{L} & \textbf{H} & \textbf{E} & \textbf{TT} & \textbf{LO} & \textbf{TP} & \textbf{CL} & \textbf{L} & \textbf{H} \\
\midrule
\midrule

\multicolumn{15}{c}{\textbf{Qwen3-VL-8B-Instruct As Generator}} \\
\midrule
\midrule
Vanilla      & 77.8 & 63.2 & 73.6 & 75.7 & 52.1 & 66.2 & 64.4 & 76.5 & 41.5 & 62.8 & 54.3 & 37.9 & 49.2 & 48.7 \\
Best-of-N    & 79.0 & 64.2 & 74.2 & 76.3 & 53.0 & 66.9 & 65.2 & 77.3 & 42.8 & 63.8 & 55.5 & 39.0 & 50.3 & 49.5 \\
Self-debug   & 79.8 & 64.8 & 74.8 & 76.8 & 53.8 & 67.6 & 65.8 & 78.0 & 43.5 & 64.5 & 56.2 & 39.8 & 51.0 & 50.2 \\
Multi-agent  & 83.0 & 70.0 & 80.5 & 74.0 & 56.8 & 70.3 & 67.2 & 86.4 & 49.5 & 76.5 & 60.8 & 44.0 & 57.7 & 57.8 \\
Qwen3-VL-2B  & 85.6 & 69.0 & 81.5 & 74.1 & 55.3 & 70.0 & 65.7 & 92.4 & 47.9 & 78.5 & 59.3 & 43.3 & 57.2 & 58.1 \\
Qwen3-VL-32B & 83.6 & 70.7 & 81.1 & 74.5 & 57.2 & 70.9 & 67.5 & 88.6 & 50.4 & 77.3 & 61.5 & 44.8 & 58.5 & 58.3 \\
Ours-SFT-2B  & 86.3 & 71.0 & 83.6 & 76.0 & 58.0 & 72.1 & 69.2 & \textbf{96.2} & 50.9 & 78.6 & 60.8 & 43.8 & 58.5 & 59.0 \\
\rowcolor{blue!15} Ours-RL-2B   & \textbf{90.0} & \textbf{72.4} & \textbf{85.3} & \textbf{77.2} & \textbf{58.8} & \textbf{73.4} & \textbf{71.8} & 93.2 & \textbf{51.6} & \textbf{79.6} & \textbf{63.5} & \textbf{45.3} & \textbf{60.0} & \textbf{60.4} \\
\midrule
\midrule

\multicolumn{15}{c}{\textbf{Qwen3-VL-32B-Instruct As Generator}} \\
\midrule
\midrule
Vanilla      & 75.5 & 66.7 & 73.9 & 76.0 & 52.5 & 67.3 & 65.9 & 78.6 & 47.8 & 69.7 & 59.3 & 46.4 & 55.8 & 56.3 \\
Best-of-N    & 77.0 & 68.0 & 75.0 & 76.2 & 53.0 & 68.1 & 67.0 & 79.5 & 48.8 & 70.2 & 60.0 & 46.8 & 56.5 & 56.8 \\
Self-debug   & 77.8 & 68.6 & 75.6 & 76.5 & 53.5 & 68.6 & 67.5 & 79.5 & 49.3 & 70.8 & 60.5 & 47.1 & 56.9 & 57.0 \\
Multi-agent  & 79.6 & 72.0 & 79.2 & 76.3 & 54.0 & 70.4 & 69.8 & 80.3 & 50.2 & 66.5 & 61.2 & 47.2 & 56.3 & 57.2 \\
Qwen3-VL-2B  & 82.5 & 72.8 & 81.0 & 76.9 & 55.7 & 71.6 & 70.6 & 83.3 & 52.5 & 72.4 & 62.6 & 49.3 & 59.2 & 59.8 \\
Qwen3-VL-32B & 80.3 & 72.7 & 79.9 & 76.6 & 54.3 & 70.9 & 70.2 & 75.0 & 50.7 & 66.8 & 61.8 & 47.5 & 56.7 & 57.5 \\
Ours-SFT-2B  & 84.8 & 74.5 & 83.3 & 77.8 & 57.6 & 73.3 & 73.6 & 87.9 & 55.4 & 76.0 & 64.7 & 52.3 & 62.1 & 61.7 \\
\rowcolor{blue!15} Ours-RL-2B & \textbf{86.7} & \textbf{76.5} & \textbf{85.5} & \textbf{78.6} & \textbf{59.8} & \textbf{75.1} & \textbf{75.9} & \textbf{90.9} & \textbf{57.0} & \textbf{78.7} & \textbf{65.9} & \textbf{54.2} & \textbf{64.0} & \textbf{63.4} \\
\midrule
\midrule

\multicolumn{15}{c}{\textbf{GPT-4o As Generator}} \\
\midrule
\midrule
Vanilla      & 92.4 & 78.1 & 92.7 & 79.9 & 75.8 & 81.6 & 82.8 & 89.4 & 57.0 & 78.3 & 71.1 & 57.8 & 66.1 & 66.6 \\
Best-of-N    & 92.8 & 79.5 & 93.5 & 81.0 & 76.8 & 82.7 & 83.5 & 90.9 & 58.5 & 79.5 & 72.5 & 58.8 & 67.3 & 67.5 \\
Self-debug   & 93.0 & 80.2 & 94.0 & 81.8 & 77.5 & 83.4 & 84.0 & 90.9 & 59.2 & 80.2 & 73.2 & 59.5 & 68.0 & 68.2 \\
Multi-agent \citep{li-etal-2025-metal} & -- & 86.3 & 95.5 & 84.2 & 79.9 & 86.5 & -- & -- & -- & -- & -- & -- & -- & -- \\
Qwen3-VL-2B  & 92.7 & 82.3 & 93.8 & 82.5 & 77.0 & 83.9 & 83.8 & 91.7 & 59.3 & 79.7 & 72.9 & 60.1 & 68.0 & 68.3 \\
Qwen3-VL-32B & 92.9 & 84.6 & 94.5 & 83.2 & 78.1 & 85.1 & 84.7 & 92.4 & 60.8 & 81.0 & 74.2 & 61.6 & 69.4 & 69.5 \\
Ours-SFT-2B  & 93.1 & 87.1 & 95.0 & 84.3 & 79.6 & 86.5 & 85.8 & 92.4 & 61.9 & 82.2 & 75.6 & 62.7 & 70.6 & 71.1 \\
\rowcolor{blue!15} Ours-RL-2B & \textbf{93.3} & \textbf{88.8} & \textbf{95.5} & \textbf{85.1} & \textbf{81.0} & \textbf{87.6} & \textbf{86.6} & \textbf{93.2} & \textbf{62.8} & \textbf{83.4} & \textbf{76.7} & \textbf{64.3} & \textbf{71.8} & \textbf{73.1} \\

\bottomrule
\end{tabular}
}
\label{tab:main_res}
\end{table*}

\subsection{Experiment Setting}
\paragraph{Datasets.}
We sample 1,000 examples from Chart2Code-160k~\citep{zhao-etal-2025-chartcoder} to build our visual-code diagnostic verification dataset that consists of 12,000 examples.
ChartMimic~\citep{yang2025chartmimic} and Plot2Code~\citep{wu-etal-2025-plot2code} are used as test datasets to verify our visual-code diagnostic verifier and test-time plot-to-code refinement.

\paragraph{Evaluation.}
The code execution success rate (E) is the basic evaluation metric.
Following \citep{yang2025chartmimic}, we calculate Text score (TT), Layout score (LO), Type score (TP), Color score (CL) and their average score (L) as fine-grained visual metrics, which are the F1 scores of these elements between the generated and the reference plot.
Besides, we employ GPT-4o to evaluate high-level (H) scores~\citep{yang2025chartmimic}.

\paragraph{Baselines.}
To evaluate the generalization of test-time refinement methods, we conduct experiments across multiple generators of different scales and architectures, including Qwen3-VL (8B/32B) and GPT-4o. This setup allows us to assess whether test-time refinement strategies can consistently improve performance across diverse plot-to-code generators.

We compare our method with several representative baselines, including vanilla one-pass generation, multi-agent cooperative generation, and representative test-time refinement methods: Best-of-N sampling and self-debugging. Best-of-N selects the best output from multiple candidates, while self-debug iteratively refines outputs using self-feedback.

For multi-agent methods, we follow the standard setting in \citep{li-etal-2025-metal}. Under the GPT-4o setting, we directly report the original results from \citep{li-etal-2025-metal}. For other generators, we reproduce the multi-agent framework using the same backbone model as the generator, where all agents share the same base model, ensuring a fair and controlled comparison.

We further include our methods, Ours-SFT-2B and Ours-RL-2B, where a compact 2B visual-code verifier is used to guide test-time refinement. To evaluate the effectiveness of learned diagnostic verification, we also compare with untrained verifiers of different scales (2B and 32B), highlighting the importance of training for accurate visual-code diagnostics.

\subsection{Main Results}

\tableautorefname \ref{tab:main_res} reports the main results across different generators and test-time scaling methods.

\paragraph{Overall improvements across generators.}
Our method consistently improves over vanilla generation across all generators and benchmarks. Both Ours-SFT-2B and Ours-RL-2B achieve notable gains on all evaluation metrics, demonstrating that the test-time refinement approach via structured visual-code diagnostics effectively corrects generation errors.

\paragraph{Comparison with test-time scaling baselines.}
Compared with representative test-time scaling methods such as Best-of-N and Self-debug, our approach achieves stronger and more consistent improvements. While these methods rely on heuristic sampling or implicit self-feedback, our method provides explicit diagnostic signals, leading to more stable refinement behavior across different generators.
The results verify the effectiveness of the proposed simple test-time refinement strategy based on visual-code diagnostics.

\paragraph{Comparison with multi-agent methods.}
Multi-agent approaches improve performance by coordinating multiple large models (e.g., Qwen3-VL-8B/32B or GPT-4o) for iterative generation and critique. In contrast, our method uses a single compact 2B verifier to guide refinement. Despite this significantly smaller model size, Ours-RL-2B achieves competitive or superior performance across most settings, indicating that learned visual-code diagnostics can effectively replace complex multi-agent reasoning for error correction.

\paragraph{Effect of verifier training.}
Comparing Ours-SFT-2B and Ours-RL-2B, we observe that reinforcement learning further improves diagnostic quality and downstream generation performance. This highlights the importance of optimizing diagnostic signals beyond supervised learning.

\paragraph{Generalization across model scales.}
Our method demonstrates consistent gains when applied to generators of different scales, from 8B to 32B models and GPT-4o. This suggests that diagnostic verification captures general error patterns that are complementary to model scaling, and can be effectively integrated with diverse base generators.

\paragraph{Inference Token Cost.}
\begin{wraptable}{r}{0.55\linewidth}
\centering
\caption{Comparison of inference token consumption.}
\label{tab:infer_token}
\resizebox{\linewidth}{!}{
\begin{tabular}{l|lcrr}
\toprule
\textbf{Method} & \textbf{Model} & \textbf{L} & \textbf{Tokens} & \textbf{Times} \\
\midrule
Best-of-N                 & GPT-4o      & 82.7 &  3047.6 & 3.7x \\
Multi-agent               & GPT-4o      & 86.5 & 11264.0 & 1.0x \\
\rowcolor{blue!15} Ours   & Qwen3-VL-2B & \textbf{87.6} & \textbf{1859.3}  & \textbf{6.1x} \\
\bottomrule
\end{tabular}
}
\end{wraptable}

\tableautorefname~\ref{tab:infer_token} compares the inference token consumption of different refinement strategies. 
Compared with Best-of-$N$ sampling and multi-agent refinement based on GPT-4o, our method uses a lightweight Qwen3-VL-2B diagnostic verifier to guide localized refinement.

Despite using substantially fewer tokens, our method achieves the best visual score ($L$: 87.6). 
Compared with multi-agent refinement, our approach reduces inference tokens from 11,264.0 to 1,859.3, corresponding to a $6.1\times$ reduction in inference cost. 
These results show that structured visual-code diagnostics enable efficient and effective test-time refinement.

\subsection{Analyses}

\paragraph{Performances over Refinement Iterations.}
\begin{figure}[t]
    \centering
\begin{subfigure}[b]{0.495\textwidth}
    \centering
\begin{tikzpicture}
\begin{axis}[
    xmin=-0.2,
    xmax=5.2,
    ymin=76.5,
    ymax=91.0,
    width=7.8cm,
    height=4.0cm,
    ytick={78, 83, 88},
    yticklabels={78, 83, 88},
    yticklabel style={font=\footnotesize, xshift=0.1cm},
    xtick = {0,1,2,3,4,5},
    xticklabels={0, 1, 2, 3, 4, 5},
    x tick label style = {font=\small, yshift=0.0cm},
    legend style={at={(0.5, 1.228)}, anchor=north, legend columns=4, font=\footnotesize, column sep=-1.0pt},
    xmajorgrids=true,
    ymajorgrids=true,
    x grid style={dashed, },
    y grid style={dashed, },
]
    \addplot[thick, solid, o_d, mark=o, mark size = 2.5pt, mark options={fill=o_d, solid}] coordinates {
        (0, 77.8) (1, 82.3) (2, 85.0) (3, 85.3) (4, 85.2) (5, 85.4)
    };\addlegendentry{Qwen-2B}
    \addplot[thick, solid, g_d, mark=diamond, mark size = 3.5pt, mark options={fill=g_d, solid}] coordinates {
        (0, 77.8) (1, 80.8) (2, 81.8) (3, 82.5) (4, 83.2) (5, 83.6)
    };\addlegendentry{Qwen-32B}
    \addplot[thick, solid, b_d, mark=triangle*, mark size = 3pt, mark options={fill=b_d, solid}] coordinates {
        (0, 77.8) (1, 81.0) (2, 83.1) (3, 84.4) (4, 85.8) (5, 86.3)
    };\addlegendentry{SFT}
    \addplot[thick, solid, r_d, mark=square*, mark size = 2pt, mark options={fill=r_d, solid}] coordinates {
        (0, 77.8) (1, 84.9) (2, 86.9) (3, 88.3) (4, 88.8) (5, 89.5)
    };\addlegendentry{RL}

\end{axis}

\end{tikzpicture}
    \caption{E score over refinement iterations.}
    \label{fig:e_iter}
\end{subfigure}
\hfill
\begin{subfigure}[b]{0.495\textwidth}
    \centering
\begin{tikzpicture}
\begin{axis}[
    ybar,
    ymax=76,
    ymin=64,
    xmax=5.5,
    xmin=-0.5,
    bar width=6,
    ytick={66, 70, 74},
    yticklabels={66, 70, 74},
    y tick label style = {font=\footnotesize, xshift=0.1cm},
    xtick = {0,1,2,3,4,5},
    xtick style = {draw=none},
    xticklabels={0, 1, 2, 3, 4, 5},
    x tick label style = {font=\small, yshift=0.13cm},
    width=7.8cm,
    height=4.0cm,
    legend style={at={(0.5, 1.228)}, anchor=north, legend columns=4, font=\footnotesize, column sep=3.8pt},
    ymajorgrids=true,
    y grid style={dashed, },
]
    \addplot[o_d, fill=o_d, bar shift=0pt, pattern=north east lines, pattern color=o_d] coordinates{
        (-0.34, 66.1)  (0.66, 69.0)  (1.66, 68.9) (2.66, 69.9)  (3.66, 69.8) (4.66, 69.9)
    };

    \addplot[g_d, fill=g_d, bar shift=0pt, pattern=crosshatch dots, pattern color=g_d] coordinates{
        (-0.115, 66.1) (0.885, 69.5)  (1.885, 70.3) (2.885, 70.1)  (3.885, 70.5)  (4.885, 70.8)
    };

    \addplot[b_d, fill=b_d, bar shift=0pt, pattern=north west lines, pattern color=b_d] coordinates{
        (0.115, 66.1) (1.115, 70.1)  (2.115, 70.8) (3.115, 71.1)  (4.115, 72.0)  (5.115, 72.15)
    };

    \addplot[r_d, fill=r_d, bar shift=0pt, pattern color=r_d] coordinates{
        (0.34, 66.1) (1.34, 71.5)  (2.34, 72.5) (3.34, 73.5)  (4.34, 74.8)  (5.34, 75.15)
    };

    \legend{Qwen-2B, Qwen-32B, SFT, RL};

\end{axis}

\end{tikzpicture}
    \caption{L score over refinement iterations.}
    \label{fig:l_iter}
\end{subfigure}
\caption{Plot-to-code generation results of refinement iterations across different visual-code diagnostic verifiers.}
\label{fig:iteration}
\end{figure}
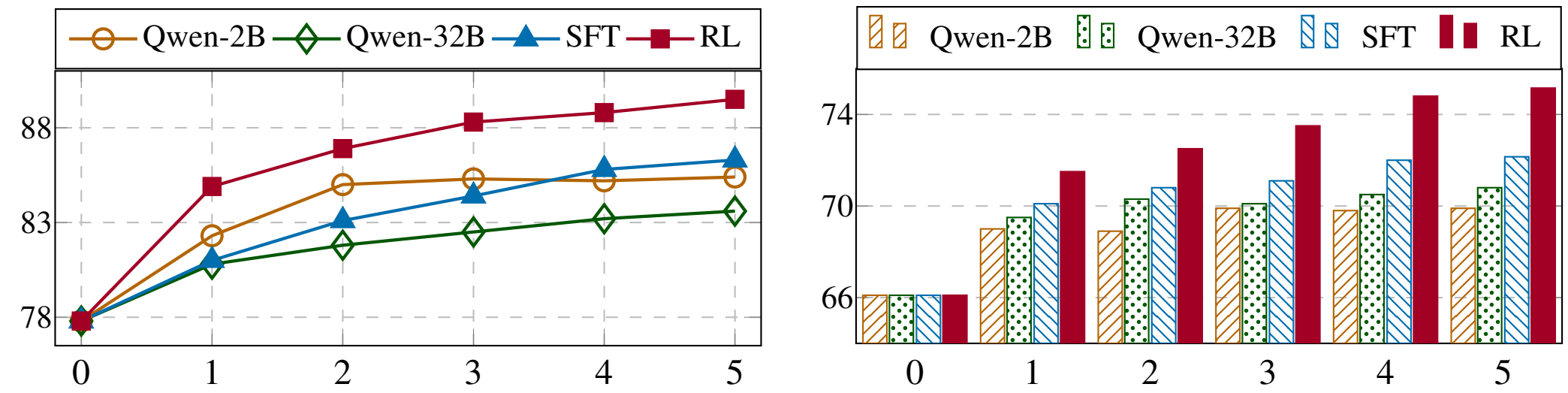
We analyze the effectiveness of visual-code diagnostics across refinement iterations on ChartMimic. 
\figureautorefname~\ref{fig:iteration} compares untrained models (Qwen-2B, Qwen-32B) and our trained verifiers (SFT and RL). Curves denote execution success rate (E), while bars report the visual metric (L).
Starting from one-pass generation (iteration 0), all methods improve with iterative refinement, confirming that test-time refinement can progressively correct both execution and visual errors. However, the quality of diagnostic signals plays a critical role. While larger untrained models such as Qwen-32B provide moderate gains, both trained 2B-scale verifiers consistently achieve stronger refinement performance, showing that learned visual-code diagnostics are more effective than model scale alone.

We further observe different refinement behaviors across verifiers. Qwen-2B achieves rapid early gains in execution success rate but provides limited improvement in visual quality, suggesting that it mainly resolves simple execution issues while struggling with more complex visual discrepancies. In contrast, trained verifiers produce more informative diagnostic feedback, enabling progressively better visual refinement over iterations.
Among all methods, the RL-trained verifier achieves the best execution and visual performance, indicating that reinforcement learning further improves diagnostic quality and refinement effectiveness.

\paragraph{Analysis on Specialized Plot-to-Code Generators.}
\begin{table*}[th]
\centering

\caption{Main results of plot-to-code generation based on ChartMaster-7B \citep{tan2025chartmaster} as Generator.}
\resizebox{1\textwidth}{!}{
\begin{tabular}{l|ccccccc|ccccccc}

\toprule
\multirow{2}{*}{\textbf{Models}} & \multicolumn{7}{c|}{\textbf{ChartMimic}} & \multicolumn{7}{c}{\textbf{Plot2Code}} \\ \cmidrule{2-15} 
& \textbf{E} & \textbf{TT} & \textbf{LO} & \textbf{TP} & \textbf{CL} & \textbf{L} & \textbf{H} & \textbf{E} & \textbf{TT} & \textbf{LO} & \textbf{TP} & \textbf{CL} & \textbf{L} & \textbf{H} \\
\midrule

\rowcolor{blue!15} Vanilla      & 93.7 & \textbf{80.0} & 91.6 & 79.1 & \textbf{67.0} & 79.4 & \textbf{85.1} & 90.1 & \textbf{57.2} & \textbf{82.4} & \textbf{62.9} & 51.0 & \textbf{63.4} & \textbf{64.5}  \\
Best-of-N    & \textbf{94.2} & 79.4 & \textbf{92.2} & 79.8 & 66.5 & \textbf{79.5} & 84.7 & \textbf{91.7} & 56.8 & 81.5 & 63.5 & 50.8 & 63.2 & 64.0 \\
Self-debug   & 91.0 & 77.0 & 89.5 & 77.5 & 64.0 & 77.0 & 82.5 & 86.4 & 55.0 & 79.0 & 61.5 & 49.5 & 61.3 & 62.0 \\
Multi-agent  & 88.5 & 74.5 & 86.5 & 75.5 & 62.5 & 74.8 & 80.5 & 84.8 & 53.0 & 76.5 & 60.0 & 48.5 & 59.5 & 60.5 \\
Qwen3-VL-2B  & 78.3 & 63.3 & 74.3 & \textbf{80.0} & 58.9 & 69.1 & 71.2 & 83.3 & 48.3 & 71.6 & 61.9 & \textbf{52.9} & 58.7 & 60.3 \\
Qwen3-VL-32B & 67.0 & 59.9 & 70.9 & \textbf{80.0} & 56.7 & 66.9 & 69.9 & 66.6 & 43.7 & 70.2 & 62.5 & 52.5 & 57.2 & 58.4 \\
Ours-SFT-2B  & 73.3 & 58.0 & 69.6 & 76.0 & 54.1 & 64.4 & 67.6 & 82.6 & 50.2 & 73.2 & 60.0 & 52.0 & 58.9 & 59.9 \\
Ours-RL-2B   & 71.9 & 56.6 & 67.9 & 76.2 & 52.9 & 63.4 & 65.5 & 87.1 & 51.2 & 70.0 & 60.7 & 53.5 & 58.9 & 59.3 \\

\bottomrule
\end{tabular}
}
\label{tab:chartmaster_res}
\end{table*}

We further evaluate ChartMaster-7B \citep{tan2025chartmaster}, a model specialized for direct plot-to-code generation.
As shown in \tableautorefname~\ref{tab:chartmaster_res}, strong one-pass generation does not necessarily translate into effective refinement.
On ChartMimic, Ours-RL-2B reduces execution success from 93.7 to 71.9 and visual similarity (L) from 79.4 to 63.4; performance also declines on Plot2Code.
This failure extends across different feedback strategies: self-debug, multi-agent refinement, and untrained verifiers also degrade performance, while Best-of-N provides almost no visual improvement.

These results suggest that the generator's ability to revise code is a key bottleneck.
ChartMaster-7B is trained for direct generation but is not explicitly trained for instruction-following code revision, whereas general-purpose MLLMs can use their broader instruction-following capabilities to act on diagnostic feedback.
To further investigate this explanation, we conduct a manual case analysis in which ChartMaster-7B receives human-annotated gold visual-code diagnostics specifying the correct error types and erroneous code regions.
Even with this accurate feedback, the model still fails to correct the identified errors in the inspected cases.
This qualitative evidence indicates that inaccurate diagnostics alone cannot explain the failures and supports insufficient code-revision ability as an important contributing factor.

ChartMaster-7B therefore illustrates an applicability boundary: effective test-time refinement requires a generator that can reliably translate diagnostic feedback into correct code edits.
This behavior suggests a trade-off in ChartMaster-7B's domain-specific fine-tuning: gains in one-pass plot-to-code generation may come at the cost of general instruction-following and code-editing capabilities.
Improving verifier quality alone may be insufficient when this capability is lacking, even for a generator with strong one-pass performance.
Training generators to follow diagnostic instructions and perform reliable code revision is an important direction for future work.

\paragraph{Ablation Study on Visual-Code Diagnostic Signals.}
\begin{wraptable}{r}{0.6\linewidth}
\centering
\caption{Ablation experiments for visual-code diagnostics.}
\label{tab:ablation_res}
\resizebox{\linewidth}{!}{
\begin{tabular}{cc|ccccccc}
\toprule
\textbf{Visual} & \textbf{Code} & \textbf{E} & \textbf{TT} & \textbf{LO} & \textbf{TP} & \textbf{CL} & \textbf{L} & \textbf{H} \\
\midrule
\ding{51} & \ding{55} & 84.2 & 69.9 & 80.1 & 75.7 & 56.9 & 70.7 & 66.6 \\
\ding{55} &	\ding{51} & 86.5 & 71.3 & 83.4 & 76.0 & 58.0 & 72.2 & 70.4 \\
\rowcolor{blue!15} \ding{51} & \ding{51} & \textbf{90.0} & \textbf{72.4} & \textbf{85.3} & \textbf{77.2} & \textbf{58.8} & \textbf{73.4} & \textbf{71.8} \\
\bottomrule
\end{tabular}
}
\end{wraptable}
We conduct ablation experiments on the ChartMimic dataset using Qwen3-VL-8B as the generator to analyze the contribution of visual and code diagnostics. As shown in \tableautorefname~\ref{tab:ablation_res}, both visual-only and code-only settings improve over vanilla generation, indicating that each type of signal provides useful guidance for refinement.

Comparing the two, code localization leads to stronger gains than visual-only signals, suggesting that precise, code-grounded feedback is particularly important for correcting errors. However, the best performance is achieved when combining both signals.
This demonstrates that visual and code diagnostics are complementary: visual signals identify what is wrong, while code localization specifies where to fix. Their joint modeling enables more accurate and effective refinement, supporting our formulation of structured visual-code diagnostics.

\paragraph{Ablation Study on Reward Function.}
\begin{wraptable}{r}{0.35\linewidth}
\centering
\small

\caption{Ablation experiments for reward functions.}
\label{tab:reward_ablation}
\begin{tabular}{l|cc}
\toprule
\textbf{Reward function} & \textbf{E} & \textbf{L} \\
\midrule
Exact match & 87.2 & 71.9 \\
F1 & 89.8 & \textbf{73.4} \\
\rowcolor{blue!15} Jaccard & \textbf{90.0} & \textbf{73.4} \\
\bottomrule
\end{tabular}
\end{wraptable}

We examine different reward functions by training verifiers from the same SFT checkpoint using exact match, set F1, or Jaccard similarity for both the visual and code rewards, while retaining their uniform weighting.
\tableautorefname~\ref{tab:reward_ablation} reports downstream execution success (E) and fine-grained visual similarity (L) on ChartMimic using Qwen3-VL-8B as the generator.

Jaccard similarity provides partial credit for overlap between variable-size sets of error types or localization actions, while normalizing for set size and penalizing both missing and spurious predictions.
In contrast, exact match assigns zero reward to a component whenever its predicted set differs from the reference, even if some predictions are correct.
Set F1 is a monotonic transformation of Jaccard similarity for each prediction/reference pair and captures similar overlap information.

Jaccard and F1 achieve the same L score of 73.4 and nearly identical execution success rates (90.0 and 89.8, respectively), suggesting limited sensitivity to the choice between these two overlap metrics in this setting.
Exact match yields lower scores (E: 87.2; L: 71.9), consistent with its sparser reward discarding useful partial matches and providing less informative feedback during training.

\paragraph{Diagnostic Verification Performance.}
\begin{wraptable}{r}{0.45\linewidth}
\centering
\caption{Visual-code diagnostic verification results across different verifiers.}
\label{tab:verifier_res}
\resizebox{\linewidth}{!}{
\begin{tabular}{l|ccc c}
\toprule
\textbf{Model} & \textbf{PEF} & \textbf{CEP} & \textbf{CER} & \textbf{CEF} \\
\midrule
Qwen3-VL-2B  & 70.5 & 50.2 & 20.1 & 28.7  \\
Qwen3-VL-32B & 74.7 & 59.4 & 40.9 & 48.4  \\
Ours-SFT-2B  & 80.8 & 65.8 & 62.1 & 63.9  \\
\rowcolor{blue!15} Ours-RL-2B   & \textbf{87.2} & \textbf{73.2} & \textbf{80.7} & \textbf{76.8}  \\
\bottomrule
\end{tabular}
}
\end{wraptable}

In addition to evaluating end-to-end plot-to-code generation, we directly assess the quality of visual-code diagnostic verification on the error discrimination dataset. We use 6,000 and 5,000 samples for SFT and RL training, respectively, and reserve 1,000 samples for evaluation.
Table~\ref{tab:verifier_res} reports the results across different verifiers. We evaluate both visual error prediction (PEF) and code localization quality (CEP, CER, CEF), where code errors are defined based on the structured tuples $(i, a, j)$ described in Section~\ref{sec:method:training}.

Our trained verifiers significantly outperform untrained models of both smaller (Qwen3-2B) and larger (Qwen3-32B) scales. Notably, Ours-RL-2B achieves the best performance across all metrics, demonstrating that a compact model can effectively learn structured visual-code diagnostics when trained with our dataset and objectives.
We further observe that untrained models exhibit imbalanced precision and recall in code localization, indicating unreliable and incomplete diagnostic signals. In contrast, our trained models achieve more balanced performance, suggesting more accurate and comprehensive identification of error regions.
These results validate the effectiveness of our diagnostic dataset and training strategy, and explain the strong gains observed in test-time refinement: improved diagnostic signals directly translate into more precise and effective code updates.

\paragraph{Case Study.}

\begin{figure*}[ht]
    \centering
    \includegraphics[width=\textwidth]{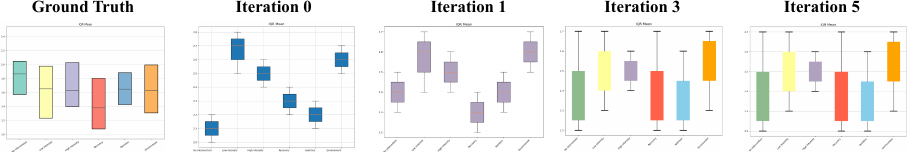}
    \caption{A case of our test-time refinement for plot-to-code generation.}
    \label{fig:casestudy}
\end{figure*}

\figureautorefname~\ref{fig:casestudy} illustrates a representative example of our visual-code diagnostics-guided test-time refinement. Iteration 0 corresponds to direct generation, which exhibits noticeable discrepancies compared to the reference chart.
With iterative refinement, the generated plot progressively aligns with the reference. Early iterations correct prominent errors such as axis configuration and styling attributes, while later iterations refine more detailed visual elements. This demonstrates that the verifier provides structured diagnostic signals that enable the generator to perform targeted updates rather than full regeneration.

We observe that some corrections may not succeed in early iterations but are resolved in subsequent steps, indicating that refinement benefits from accumulating diagnostic feedback over iterations. Meanwhile, later iterations tend to introduce fewer changes, suggesting convergence as major errors are resolved.
Despite these improvements, certain fine-grained discrepancies remain (e.g., subtle styling differences), even after multiple iterations. 
This highlights the remaining challenges in precisely aligning low-level visual details and suggests that further improving diagnostic granularity or generator editing capability could yield additional gains.

\section{Conclusion}

We study plot-to-code generation from the perspective of test-time refinement and identify diagnostic verification as a key component for improving generation quality.
We propose a visual-code diagnostic framework that produces structured signals to detect visual discrepancies and localize code errors.
By constructing a scalable diagnostic dataset and training a compact verifier, our method enables effective and targeted code refinement.
Experiments show that a 2B verifier consistently improves generation performance, outperforming larger untrained models and achieving competitive results compared to multi-agent approaches with substantially lower cost.
We further analyze limitations on specialized plot-to-code generators, highlighting the importance of revision capability for effective refinement.
Our results suggest that learned diagnostic verification is a promising direction for efficient multimodal code generation.

\begin{ack}
This work was supported by the National Natural Science Foundation of China under Grant 62336008, the Guangdong Provincial Key Laboratory of Artificial General Intelligence and Large Models under Grant 2026B1212010005, and the CCF-Huawei Populus Grove Fund under Grant CCF-HuaweiSE2025021.
\end{ack}

{
\bibliographystyle{unsrtnat}
\bibliography{custom}
}


\appendix


\appendix
\label{sec:appendix}

\section{Plot Error Taxonomy and Distribution}
\label{sec:app:error_taxonomy}

\subsection{Plot Error Types}
Examining all results and considering the programming logic and coding style of the Matplotlib~\citep{Hunter:2007,harris2020array} code, we categorize generation errors into four coarse-grained types:
\begin{itemize}
    \item Compiler error, where the generated code fails to execute and render the plot.
    \item Data error, where the number and values of shapes and data points are incorrect.
    \item Local style error, where styles of specific graphic shapes and elements are wrong, such as incorrect colors, markers, or line styles.
    \item Global style error, where styles of global elements for the plot are misaligned, like axes, grid, titles, or extra texts.
\end{itemize}

\begin{wrapfigure}{r}{0.5\linewidth}
\centering
\includegraphics[width=0.5\textwidth]{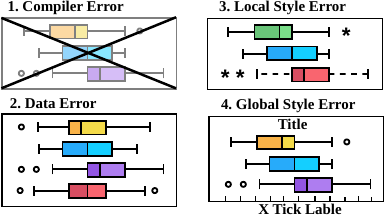}
\caption{Error types for plot-to-code generation.}
\label{fig:error}
\end{wrapfigure}

\figureautorefname~\ref{fig:error} demonstrates examples of four error types.
The compiler error does not render the reference plot image, while the data error contains an extra box and incorrect lower and upper quartiles.
There are wrong shapes and colors for the lines, dots, and boxes in the local style error, while extra x tick label and title in the global style error.
In particular, the categorization basis of the latter three error types is primarily derived from the Matplotlib~\citep{Hunter:2007} code.
A standard Matplotlib program consists of three components.
The first component is data loading, the second involves invoking the core plotting APIs, where the local properties of plot elements can be configured, and the third sets global elements such as the title, axes, and additional annotations.
Accordingly, these three categories of plot errors directly correspond to errors in the code, enabling efficient localization of code errors from plot errors.
This correspondence reduces the burden on the subsequent Plot\&Code discriminator, allowing even relatively small-scale MLLMs to effectively handle this task and thereby improve the iterative refinement performance of the plot-to-code generator.

\subsection{Plot Error Distribution}
\begin{figure*}[th]
\centering

\begin{tikzpicture}

\begin{axis}[
    xbar stacked,
    xmin=0,
    xmax=100,
    ymin=-0.7,
    ymax=3.7,
    width=13cm,
    height=4.5cm,
    bar width=12,
    ytick = {0, 1, 2, 3},
    yticklabels={\textbf{GPT-4o}, \textbf{Qwen3-VL-32B}, \textbf{Qwen3-VL-8B}, \textbf{ChartMaster-7B}},
    yticklabel style={font=\normalsize, xshift=0.0cm, rotate=0},
    xtick={0, 25, 50, 75, 100},
    xtick style={draw=none},
    xticklabels={},
    xticklabel style={font=\footnotesize, xshift=0.0cm, yshift=0.2cm, rotate=0},
    legend style={at={(0.5, 1.2)}, anchor=north, legend columns=4, font=\normalsize, draw=none},
    axis line style={draw=none},
    after end axis/.code={
        \node[font=\normalsize, text=r_d] at (axis cs:5,0) {10.1};
        \node[font=\normalsize, text=r_d] at (axis cs:11.0,1) {21.9};
        \node[font=\normalsize, text=r_d] at (axis cs:10,2) {19.5};
        \node[font=\normalsize, text=r_d] at (axis cs:2.2,3) {4.4};
        \node[font=\normalsize, text=b_d] at (axis cs:22.5,0) {24.8};
        \node[font=\normalsize, text=b_d] at (axis cs:30,1) {15.8};
        \node[font=\normalsize, text=b_d] at (axis cs:31,2) {22.7};
        \node[font=\normalsize, text=b_d] at (axis cs:21,3) {33.3};
        \node[font=\normalsize, text=g_d] at (axis cs:51.5,0) {33.3};
        \node[font=\normalsize, text=g_d] at (axis cs:51,1) {26.3};
        \node[font=\normalsize, text=g_d] at (axis cs:55,2) {26.6};
        \node[font=\normalsize, text=g_d] at (axis cs:53,3) {30.7};
        \node[font=\normalsize, text=o_d] at (axis cs:84,0) {31.8};
        \node[font=\normalsize, text=o_d] at (axis cs:82,1) {36.0};
        \node[font=\normalsize, text=o_d] at (axis cs:84,2) {31.2};
        \node[font=\normalsize, text=o_d] at (axis cs:84,3) {31.6};
    },
]
    \addplot[r_d, fill=r_l] coordinates {
        (10.1, 0) (21.9, 1) (19.5, 2) (4.4, 3) 
    };
    \addplot[b_d, fill=b_l] coordinates {
        (24.8, 0) (15.8, 1) (22.7, 2) (33.3, 3)
    };
    \addplot[g_d, fill=g_l] coordinates {
        (33.3, 0) (28.3, 1) (26.6, 2) (30.7, 3)
    };
    \addplot[o_d, fill=o_l] coordinates {
        (31.8, 0) (36.0, 1) (31.2, 2) (31.6, 3)
    };
    \legend{Compile Error, Data Error, Local Style Error, Global Style Error};

\end{axis}
\end{tikzpicture}

\caption{Distribution of plot error types.}
\label{fig:distribution}
\end{figure*}
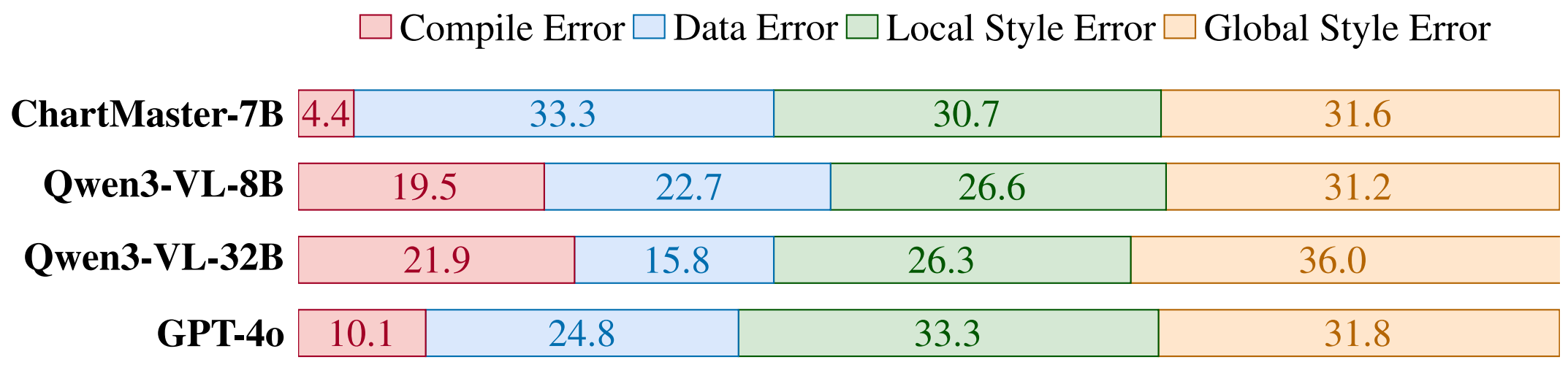

We sample 100 instances from the ChartMimic dataset and use four representative plot-to-code models to directly generate code outputs.
The three authors of this paper first independently inspect the code and rendered plots produced by the large language models, and then jointly discuss and establish the error taxonomy by synthesizing their individual judgments. 
Subsequently, all samples are annotated by the three authors.
For cases with annotation disagreements, the final labels are determined through group discussion.
\figureautorefname~\ref{fig:distribution} presents the distribution of error types across different models: Qwen3-VL-8B, Qwen3-VL-32B, ChartMaster and GPT-4o.

\section{Dataset Construction}
\label{sec:app:dataset_construction}
\subsection{Process}

\subsubsection{Error-Aware Code Mutation}

Given a correct chart-code pair $(I, C)$, we construct synthetic erroneous samples by applying controlled code mutations. We focus on three types of visual errors: data errors, local style errors, and global style errors. Compiler errors are excluded from mutation, as they can be directly identified from execution feedback.
For each seed program, we employ GPT-4o to sequentially generate three mutated variants, each introducing one type of error while preserving other parts of the code.
The prompts of GPT-4o for code mutation are listed in \appendixautorefname~\ref{sec:app:dataset:prompt}.
The order of mutation types is randomly permuted across samples to improve diversity and avoid positional bias.

\subsubsection{Pairwise Sample Construction}

Given one seed program and its three mutated variants, we construct pairwise diagnostic samples by selecting two programs $(C_1, C_2)$. The corresponding rendered plots $(I_1, I_2)$ are obtained via execution.
We treat $C_1$ as the generated code and $C_2$ as the reference. The visual error label is determined by the mutation type that differentiates $C_1$ from $C_2$. This pairwise construction produces multiple combinations per seed example, resulting in up to 12 samples per original chart-code pair.

\subsubsection{Code Region Alignment}

To obtain code localization supervision, we align $C_1$ and $C_2$ using a line-level variant of the Levenshtein distance algorithm (in \appendixautorefname~\ref{sec:app:dataset:algo}). The alignment identifies inserted, deleted, and modified code segments between the two programs.
We further refine the alignment by applying code-aware constraints to ensure valid line matching and reduce noise. The resulting regions are treated as candidate error locations.

\subsubsection{Diagnostic Annotation Format}

Instead of predicting explicit line indices, we convert localization signals into a generation-friendly format by inserting \texttt{TODO} comments near the identified code regions in $C_1$. This produces structured diagnostic annotations that preserve code context and are more robust for language models.
Each final training instance consists of: 1. Generated plot $I_1$, 2. Reference plot $I_2$, 3. Generated code $C_1$ with diagnostic annotations, 4. Visual error label.
This construction provides scalable supervision for both error classification and fine-grained code localization without requiring manual annotation of code regions.

\subsection{Postprocessing Algorithm}
\label{sec:app:dataset:algo}
\begin{algorithm*}[ht]
\caption{Adapted Line-level Levenshtein distance algorithm}
\label{alg:code-diff-lines}
\LinesNumbered

\KwIn{
Two code strings $code1,\,code2$
}
\KwOut{
$inserted$, $deleted$, $replaced$ pairs of code line numbers in $code1$ and $code2$
}

$lines1 \leftarrow \mathtt{SplitLines}(code1)$\;
$lines2 \leftarrow \mathtt{SplitLines}(code2)$\;
$n \leftarrow |lines1|,\; m \leftarrow |lines2|$\;

\textbf{Initialize DP table}\;
$dp \leftarrow \mathtt{Matrix}(n+1, m+1, 0)$\;
\For{$i \leftarrow 0$ \KwTo $n$}{
    $dp[i][0] \leftarrow i$\tcp*{delete $i$ lines}
}
\For{$j \leftarrow 0$ \KwTo $m$}{
    $dp[0][j] \leftarrow j$\tcp*{insert $j$ lines}
}

\textbf{Fill DP table}\;
\For{$i \leftarrow 1$ \KwTo $n$}{
    \For{$j \leftarrow 1$ \KwTo $m$}{
        \uIf{$lines1[i-1] = lines2[j-1]$}{
            $dp[i][j] \leftarrow dp[i-1][j-1]$
        }\uElse{
            $delete\_distance \leftarrow dp[i-1][j] + 1$\;
            $insert\_distance \leftarrow dp[i][j-1] + 1$\;
            \tcp{The following condition control for replace\_distance is different from the conventional Levenshtein distance algorithm}
            \uIf{$\texttt{common\_prefix\_length}(lines[i-1], lines2[j-1]) >= 6$}{
                $replace\_distance \leftarrow dp[i-1][j-1] + 1$\;
            }
            \uElse{
                $replace\_distance \leftarrow dp[i-1][j-1] + 999$\;
            }
            $dp[i][j] \leftarrow \min(delete\_distance, insert\_distance, replace\_distance)$\;
        }
    }
}

\textbf{Backtracking to extract operations pairs}\;
$i \leftarrow n,\; j \leftarrow m$\;
$inserted, deleted, replaced \leftarrow [\,],\, [\,],\, [\,]$\;

\While{$i > 0 \;\mathbf{or}\; j > 0$}{
    \uIf{$i>0 \; \mathbf{and} \; j>0 \; \mathbf{and} \; lines1[i-1] = lines2[j-1]$}{
        $i \leftarrow i-1,\; j \leftarrow j-1$\tcp*{match}
    }
    \uElseIf{$j>0 \; \mathbf{and} \; dp[i][j] = dp[i][j-1] + 1$}{
        $inserted.\mathtt{append}((i, j))$\;
        $j \leftarrow j-1$\tcp*{insert}
    }
    \uElseIf{$i>0 \;\mathbf{and}\; dp[i][j] = dp[i-1][j] + 1$}{
        $deleted.\mathtt{append}((i, j))$\;
        $i \leftarrow i-1$\tcp*{delete}
    }
    \uElseIf{$i>0 \; \mathbf{and} \; j>0 \; \mathbf{and} \; dp[i][j] = dp[i-1][j-1] + 1$}{
        $replaced.\mathtt{append}((i, j))$\;
        $i \leftarrow i-1,\; j \leftarrow j-1$\tcp*{replace}
    }
    \uElse{
        \KwSty{raise} $\mathtt{ValueError}\text{(``Unexpected Case'')}$\;
    }
}

\Return $inserted, deleted, replaced$\;
\end{algorithm*}

We adopt line-level Levenshtein distance algorithm to compare two codes and extract the inserted/deleted/modified code line numbers, which is detailed in \algorithmautorefname~\ref{alg:code-diff-lines}.
In concrete terms, the condition control statement in lines 20-23 for replace\_distance is different from the conventional Levenshtein distance algorithm.

\subsection{Prompts}
\label{sec:app:dataset:prompt}
\begin{figure*}[ht]
\centering

\begin{tcolorbox}[title=\textbf{Dataset Construction Prompt for Data Issue}, colbacktitle=gray!30, coltitle=black, colback=white]

You are an expert Python developer who specializes in writing matplotlib code. Please modify the given code snippet according to the requirements.

Requirements:

1. Change the data used in the code snippet to generate a different image.

2. The changed data should have enough variation to ensure the noticeable difference in the generated image.

3. Add or reduce the number of graphic shapes, and change the values of the data.

4. Keep the color and the style of graphic shapes of the image unchanged.

5. Ensure that the overall structure and style of the code remain unchanged.

Response Format:

Return the modified complete code surrounded by {\fontfamily{pcr}\selectfont\symbol{96}\symbol{96}\symbol{96}}python and {\fontfamily{pcr}\selectfont\symbol{96}\symbol{96}\symbol{96}}.

\end{tcolorbox}

\begin{tcolorbox}[title=\textbf{Dataset Construction Prompt for Local Style Issue}, colbacktitle=gray!30, coltitle=black, colback=white]

You are an expert Python developer who specializes in writing matplotlib code. Please modify the given code snippet according to the requirements.

Requirements:

1. Change the style of graphic shapes and elements in the chart to generate a different image.

2. Change, add or reduce various parameters in the graphic shape rendering function.

3. The change should have enough variation to ensure the noticeable difference in the generated image.

4. Only when the original code uses the pie function, add or change or reduce parameters wedgeprops or explode with one-half probability. 

5. Keep the data and the style of axes of the image unchanged.

6. Ensure that the overall structure and style of the code remain unchanged.

Response Format:

Return the modified complete code surrounded by {\fontfamily{pcr}\selectfont\symbol{96}\symbol{96}\symbol{96}}python and {\fontfamily{pcr}\selectfont\symbol{96}\symbol{96}\symbol{96}}.

\end{tcolorbox}

\begin{tcolorbox}[title=\textbf{Dataset Construction Prompt for Global Style Issue}, colbacktitle=gray!30, coltitle=black, colback=white]

You are an expert Python developer who specializes in writing matplotlib code. Please modify the given code snippet according to the requirements.

Requirements:

1. Change the global styles used in the code snippet to generate a different image.

2. Gloabl styles include: Axis ranges, Axis scales, Axis labels, Axis ticks, Title, Annotations, Grid, and Extra text.

3. The change should have enough variation to ensure the noticeable difference in the generated image.

4. Keep the data and the color and other styles of graphic shapes of the image unchanged

5. Ensure that the overall structure and style of the code remain unchanged.

Response Format:

Return the modified complete code surrounded by {\fontfamily{pcr}\selectfont\symbol{96}\symbol{96}\symbol{96}}python and {\fontfamily{pcr}\selectfont\symbol{96}\symbol{96}\symbol{96}}.

\end{tcolorbox}

\caption{Dataset Construction Prompts.}
\label{fig:llm_prompt}
\end{figure*}

We mutate plot code by GPT-4o based on the proposed taxonomy of plot error during dataset construction.
Prompts for code mutation are listed in \figureautorefname~\ref{fig:llm_prompt}.

\section{Test-Time Refinement for Plot-to-Code Generation via Visual-Code Diagnostics}
\label{sec:app:iterrefine}
\begin{figure*}[ht]
\centering

\begin{tcolorbox}[title=\textbf{Prompt of visual-code diagnostic verifier}, colbacktitle=gray!30, coltitle=black, colback=white]

I write a code snippet to reproduce a science chart. I will give you the reference chart and the generated chart by my code.
Please help me to modify my code to make the reference chart closer to the original chart.
Output difference types and code snippet with TODO comments.

Difference types include: 

1. data: number and values of shapes and data points

2. local\_style: styles of specific graphic shapes and elements

3. global\_style: styles of global elements like Axis, Title, Annotations, Grid, Extra text

TODO comments should be inserted into my code snippet and conform the following format:

1. \# TODO: Add extra code in the next line 

2. \# TODO: Modify the following <number> lines

3. \# TODO: Delete the following <number> lines

Return difference types surrounded < and > in one line.

Return the modified complete code surrounded by {\fontfamily{pcr}\selectfont\symbol{96}\symbol{96}\symbol{96}}python and {\fontfamily{pcr}\selectfont\symbol{96}\symbol{96}\symbol{96}}.

Reference Chart:

\{reference image\}

Generated Chart:

\{generated image\}

Code Snippet:

{\fontfamily{pcr}\selectfont\symbol{96}\symbol{96}\symbol{96}}python
 
\{generated code\}
 
{\fontfamily{pcr}\selectfont\symbol{96}\symbol{96}\symbol{96}}
\end{tcolorbox}

\caption{Prompt of visual-code diagnostic verifier.}
\label{fig:verifier_prompt}
\end{figure*}

\begin{figure*}[ht]
\centering
\begin{tcolorbox}[title=\textbf{Prompt of test-time refinement for plot-to-code generator}, colbacktitle=gray!30, coltitle=black, colback=white]

You are a helpful assistant who can revise Python matplotlib code to reproduce the reference chart image.

I will provide:

1. A Python code snippet that generates a chart, but the output does not match the reference chart image. The code contains \# TODO comments indicating the specific locations that require modification.

2. A list of difference types describing how the generated chart differs from the reference image.

Your task is:

1. Carefully analyze and infer the visual differences between the reference chart image and my code snippet.


2. Identify specific code segments that need to be modified.


3. Return only the code segments that need to be modified, not the complete code.

IMPORTANT: Return the modifications in the following format:

---OLD---

[original code segment 1]

---NEW---

[modified code segment 1]

---OLD---

[original code segment 2]

---NEW---

[modified code segment 2]

You can have multiple OLD/NEW pairs if multiple segments need to be modified.

Code snippet:

{\fontfamily{pcr}\selectfont\symbol{96}\symbol{96}\symbol{96}}python

\{current code\}

{\fontfamily{pcr}\selectfont\symbol{96}\symbol{96}\symbol{96}}

Difference types:

\{Plot error types\}

Reference image:

\{reference image\}

Generated image:

\{generated image\}

\end{tcolorbox}

\caption{Prompts of test-time refinement for plot-to-code generation via visual-code diagnostics.}
\label{fig:refinement_prompt}
\end{figure*}

During iterative refinement, we use prompts in \figureautorefname~\ref{fig:verifier_prompt} to instruct the visual-code diagnostic verifier and \figureautorefname~\ref{fig:refinement_prompt} for the plot-to-code generator, respectively.

\section{Experiments}

\subsection{Hyperparameters}
We train the visual-code diagnostic verifier based on Qwen3-VL-2B-Instruct~\citep{qwen3vl}.
For the SFT stage, we conduct training on the 6,000 samples for 2 epochs.
Our implementation is based on LLaMA-Factory \citep{zheng-etal-2024-llamafactory} with batch size 64, learning rate 1e-5, cutoff length 16,000.
For the reinforcement learning stage, we continually train the SFT model on the other 5,000 samples for 3 epochs.
Our training is based on VERL~\citep{10.1145/3689031.3696075} with global batch size 32, ppo\_mini\_batchsize 16, learning rate 1e-6, rollout\_n 8, sampling temperature 1.0, top\_p 1.0.
For the inference stage, we repeat iterative refinement $N=5$ times.
We set the temperature to 0.7, top\_p 0.8 for our verifiers, while default sampling parameters for all plot-to-code generators.
All experiments are conducted on 8 Nvidia A6000 GPUs.

\subsection{Case Study}
\begin{figure*}[ht]
    \centering
    \includegraphics[width=\textwidth]{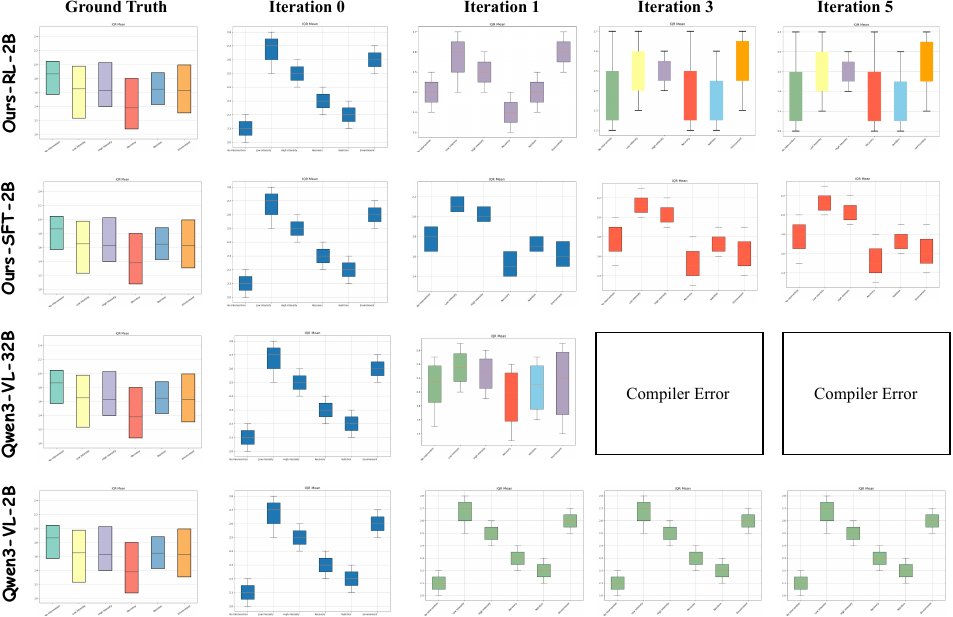}
    \caption{A case of test-time refinement for plot-to-code generation with different visual-code diagnostic verifiers.}
    \label{fig:case_appendix}
\end{figure*}

\figureautorefname~\ref{fig:case_appendix} illustrates a case of test-time refinement for plot-to-code generation with different visual-code diagnostic verifiers.

\end{document}